\documentclass[preprint,authoryear,11pt,letterpaper]{elsarticle}
\usepackage[top=0.75in, bottom=0.75in, left=0.75in, right=0.75in]{geometry}

\usepackage{amsmath,amsthm}
\usepackage{amssymb}
\usepackage{breqn}
\usepackage{amsfonts}
\usepackage{graphicx}
\usepackage{enumitem}
\usepackage{algorithmic}

\newbox{\bigpicturebox}
\graphicspath{{graphics/}}

\usepackage{xcolor}
\definecolor{darkblue}{rgb}{0.0,0.5,0.5}
\definecolor{blue}{rgb}{0.0,0.5,0.68}
\usepackage[colorlinks]{hyperref}
\hypersetup{colorlinks,breaklinks,linkcolor=darkblue,urlcolor=darkblue,anchorcolor=darkblue,citecolor=darkblue}
\usepackage{booktabs}
\usepackage{multirow}
\usepackage{lscape}
\usepackage{caption}
\usepackage{epstopdf}
\usepackage{lineno}
\usepackage{float}
\usepackage{microtype}
\usepackage{lineno}
\usepackage{placeins}

\usepackage{algorithm}
\usepackage{amsfonts}
\usepackage{graphicx}
\usepackage{microtype}
\usepackage{booktabs}
\usepackage[colorlinks]{hyperref}
\hypersetup{colorlinks,breaklinks,linkcolor=blue,urlcolor=blue,anchorcolor=blue,citecolor=blue}
\usepackage{booktabs} 
\usepackage{mathtools}
\usepackage{color}
\usepackage{adjustbox}
\usepackage[figuresright]{rotating}
\usepackage{tablefootnote}

\usepackage{tikz}
\usepackage{hyperref}
\usepackage{threeparttable}

\usepackage{amsfonts}
\usepackage{graphicx}
\usepackage{microtype}
\usepackage{booktabs}
\usepackage[colorlinks]{hyperref}
\hypersetup{colorlinks,breaklinks,linkcolor=blue,urlcolor=blue,anchorcolor=blue,citecolor=blue}
\usepackage{booktabs} 
\usepackage{mathtools}
\usepackage{color}
\usepackage{adjustbox}

\usepackage{cleveref}
\usepackage{natbib}
\setcitestyle{square, comma, numbers,sort&compress, super}

\usepackage{tikz}
\newcommand*\emptycirc[1][1ex]{\tikz\draw (0,0) circle (#1);} 
\newcommand*\halfcirc[1][1ex]{%
  \begin{tikzpicture}
  \draw[fill] (0,0)-- (90:#1) arc (90:270:#1) -- cycle ;
  \draw (0,0) circle (#1);
  \end{tikzpicture}}
\newcommand*\fullcirc[1][1ex]{\tikz\fill (0,0) circle (#1);} 

\begin{document}
\nolinenumbers
\begin{frontmatter}

\title{Improving the generalizability and robustness of large-scale traffic signal control}

\author{Tianyu Shi}
\ead{ty.shi@mail.utoronto.ca}

\address{Department of Civil Engineering, University of Toronto, 35 St. George Street,Toronto, Ontario, M5S 1A4, Canada}

\author{François-Xavier Devailly}

\author{Denis Larocque}

\author{Laurent Charlin}

\address{Department of Decision Sciences at HEC Montreal,Quebec, Canada}

\cortext[cor1]{Corresponding author. Address: 35 St.George Street, Toronto, Ontario, M5S 1A4, Canada}

\begin{abstract}
A number of deep reinforcement-learning (RL) approaches propose to control traffic signals. Compared to traditional approaches, RL approaches can learn from higher-dimensionality input road and vehicle sensors and better adapt to varying traffic conditions resulting in reduced travel times (in simulation). However, these RL methods require training from massive traffic sensor data. To offset this relative inefficiency, some recent RL methods have the ability to first learn from small-scale networks and then generalize to unseen city-scale networks without additional retraining (\emph{zero-shot transfer}). In this work, we study the robustness of such methods along two axes. First, sensor failures and GPS occlusions create missing-data challenges and we show that recent methods remain brittle in the face of these missing data. Second, we provide a more systematic study of the generalization ability of RL methods to new networks with different traffic regimes. Again, we identify the limitations of recent approaches.
We then propose using a combination of distributional and vanilla reinforcement learning through a policy ensemble. Building upon the state-of-the-art previous model which uses a decentralized approach for large-scale traffic signal control with graph convolutional networks (GCNs), we first learn models using a distributional reinforcement learning (DisRL) approach. In particular, we use implicit quantile networks (IQN) to model the state-action return distribution with quantile regression. For traffic signal control problems, an ensemble of standard RL and DisRL yields superior performance across different scenarios, including different levels of missing sensor data and traffic flow patterns. Furthermore, the learning scheme of the resulting model can improve zero-shot transferability to different road network structures, including both synthetic networks and real-world networks (e.g., Luxembourg, Manhattan). We conduct extensive experiments to compare our approach to multi-agent reinforcement learning and traditional transportation approaches. Results show that the proposed method improves robustness and generalizability in the face of missing data, varying road networks, and traffic flows.
\end{abstract}


\end{frontmatter}


\section{Introduction}

%
%
%
%
As the number of cars on our roads continues to rise it is imperative to adapt road networks to minimize congestion. Developing robust yet efficient traffic control strategies is a powerful mitigator~\citep{wei2018intellilight,devailly2021ig,wei2019colight}. Powerful traffic signal control (TSC) methods, for example, based on deep reinforcement learning~\cite{silver2017mastering}, now exist to optimize the control signal phase (e.g., red or green). They learn from and use available historical and real-time traffic and vehicle data~\citep{shi2019driving,essa2020self,wei2019colight,varaiya2013max}.

Real-time data can be collected from the built-in sensors of the vehicles and then transmitted to the control system to help in decision-making (e.g., to free busy lanes by changing the phase of the TSC)~\citep{zhang2020using}. However, missing values in the collected data from vehicles~\citep{nanthawichit2003application}, (e.g., caused by GPS occlusions and transmission delays) --- are common. Downstream, missing data will introduce uncertainty in the observations of the system, which will then be challenging for the decision-making module. 
Controlling traffic signals under these exogenous sources of uncertainty requires robust control policies. 

A second challenge is that traffic conditions can be non-stationary because of singular events such as accidents and construction and also due to recurring patterns (e.g., periodic daily and weekly ones). They can also evolve over time as a result of other infrastructure changes (e.g., new roads nearby). As a result, it is advantageous to use control policies that can adapt to new scenarios, varying traffic-flow patterns, and even allow deployment across networks of different scales.

The ability to obtain policies that are both robust (to sensor failures) and that can generalize to new situations (traffic and networks) is important for deploying control policies in complex road systems that are ubiquitous in our cities. Current methods do not yield policies with both desiderata (we show this below). This is the gap we address in this paper. Next, we introduce the classes of existing approaches for traffic signal control.


First, hand-crafted policies for TSCs form a class of traditional approaches. For example, \emph{fixed-time} approaches~\citep{koonce2008traffic} define a fixed cycle length and phase time for each intersection based on the road configuration. \emph{\emph{Greedy}}~\citep{varaiya2013max} maximizes the throughput of the road networks by greedily picking the phase that can maximize the pressure. In principle, hand-crafted policies generalize across networks and traffic conditions. However, they rely  on unrealistic assumptions, such that the road lanes have unlimited capacity and that the traffic flow is constant. As a result, their application in real-world and complex road networks is limited~\citep{varaiya2013max}.

Reinforcement learning (RL), a formalism for sequential decision-making, is proving to be an effective tool to learn complex policies for diverse traffic-control problems~\citep{wei2018intellilight,wei2019colight,chu2019multi}. RL models traffic signals as agents that use the current \emph{state} of the environments (e.g., the position of all nearby vehicles) to control the light phase. Reinforcement learning agents are trained to maximize a utility function called a \emph{reward}. For traffic-signal control, rewards are often taken to be proxies of the traffic efficiency, measured, for example, as the inverse (vehicle) delay or queue length. In simulation, RL has been trained to control traffic lights in real-world road networks and outperforms hand-crafted policies~\citep{wei2018intellilight,koonce2008traffic}.



\begin{figure}[h]
\centering
\includegraphics[width=8cm]{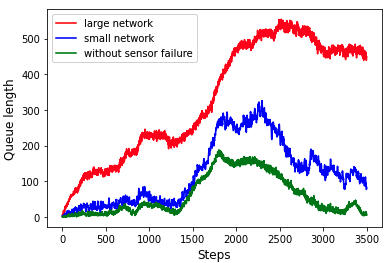}
\caption{Sensor failures can create larger delays in large networks compared to small networks. 
In the experiment, the large-scale network has 30 intersections while the small-scale network has 3 intersections. We tune the traffic demand parameter so that both small and large networks have a similar queue length. As a result, we can obtain a comparable baseline (shown in green).}
\label{sensor}
\end{figure}

RL has shown robustness in small-scale road networks (one to five intersections). In particular, the standard Deep Q-Networks (DQNs) for RL, using a replay buffer to store previous experiences, have demonstrated a level of generalizability for different traffic demands.~\citep{rodrigues2019towards,zhang2020using}.  Figure~\ref{sensor} shows that DQNs still suffer from a performance decrease when faced with missing data. The performance further decreases in larger road networks.


Generalizability is also important for RL policies since training RL agents is computationally costly even for small-scale networks.
To scale agents to larger-scale road networks (of the order of neighborhoods or whole cities) with different traffic flow patterns, \cite{wei2019colight} and \cite{devailly2021ig} explore scalable and decentralized multi-agent reinforcement learning (MARL) approaches.
In particular, to encourage better utilization of the spatial-temporal information, researchers model the road network using graph neural networks~\citep{zhou2018graph} trained with RL to encourage cooperation \citep{wei2019colight} and improve transferability \citep{ devailly2021ig}.

\begin{figure}[h]
\centering
\includegraphics[width=8cm]{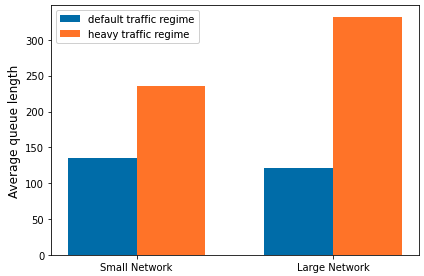}
\caption{The comparison of different road networks given different traffic demands. In the test, we tune the arrival rate to make two networks have similar congestion (i.e., average queue length across the whole simulation steps), then increase the traffic regime (density) by two times to simulate the demand surge.}
\label{demand}
\end{figure}


We are interested in further studying these approaches. In particular, we investigate their robustness to missing data as well as their ability to generalize to larger-size networks with different traffic regimes. 

We introduce an initial experiment to demonstrate the limitation of current deep-reinforcement learning approaches. We learn a traffic signal control agent based on decentralized independent deep reinforcement learning~\citep{rodrigues2019towards}. We also add a few standard Deep RL tricks: Double Q-Learning \citep{hasselt2010double} to prevent overestimation and to stabilize the learning process, and parameter noise for exploration \citep{fortunato2017noisy}. The experiment compares the performance of this Deep RL agent trained on a small network with 3 intersections and tested on the same small network as well as a larger one with 30 intersections. Sensor failures are also presented in the test scenarios (the exact setup is described later \ref{setupexp}).

As noted above, we find that faced with sensor failures, the RL agent performs comparatively worse in a large road network versus in a small one (Figure~\ref{sensor}). Furthermore, we find that when demand surges,\footnote{The heavy traffic regime is simulated by doubling the number of cars in the network.} the performance decreases more in the large road network (Figure~\ref{demand}). This result demonstrates that a shift in the distribution of network architectures and the distribution of demand hinders the robustness of reinforcement learning approaches. These observations~\ref{sensor} and~\ref{demand} motivate the development of robust and transferable Deep RL-based  methods for traffic signal control.


In this work, we propose RGLight, a method that can further improve both the robustness and generalizability of traffic-signal controllers compared to previous works (as shown in Table~\ref{survey}). RGLight uses distributional RL (DisRL)~\citep{bellemare2017distributional, dabney2018implicit}. Compared to standard RL that estimates the mean value of \emph{returns} (actions in each state), DisRL constructs a (full) distribution over returns. DisRL tends to improve the stability of the learning process, i.e., improve convergence, especially in dynamic environments~\citep{bellemare2017distributional, lyle2019comparative}. Until now, DisRL instantiations focus on the single-agent setting without exogenous uncertainty. We conjecture that DisRL can also improve the learning stability in multi-agent settings and in particular 
in large-scale traffic signal control settings. 

Building upon the prior work of IGRL~\citep{devailly2021ig}, we find that a policy ensemble that combines distributional and deterministic modeling further boosts the generalizability of IGRL across a number of scenarios.

We also propose several criteria to evaluate the robustness and generalizability of the learned policies and conduct extensive experiments to evaluate RGLight in both real-world settings and synthetic settings. Results show that RGLight improves the robustness and generalizability of traffic signal control compared to several state-of-the-art baselines.

To summarize, our main contributions are:
\begin{itemize}
    \item A method based on a policy ensemble of distributional RL and standard graph-based RL for traffic signal control. Our approach focuses on improving the overall generalization performance and robustness of the trained RL policies.
    
    
    

    \item An empirical evaluation with different types of missing values, flow patterns, and network structures using both synthetic and real-world road networks. We compare approaches using an \emph{evaluation matrix} to provide a more systematic analysis of the generalization ability of different models. We highlight that RGLight outperforms several state-of-the-art baselines.
\end{itemize}


\section{Background and Related work}
\subsection{RL-based Traffic Signal Control} The very first implementation of RL in TSC uses tabular Q-Learning to learn from a single intersection~\citep{wiering2004intelligent}. \citet{cai2009adaptive} then uses RL with function approximations. However, most previous investigations are limited to toy scenarios. To develop RL methods for more realistic traffic data, researchers turned their attention to deep RL. \citet{wei2018intellilight,shabestary2022adaptive} show that deep reinforcement learning can dynamically adjust to real-time traffic. However, the high dimension of the joint action space still limits the scalability of centralized RL approaches. 

\subsection{Large-Scale Traffic Signal Control} Multi-agent Reinforcement Learning (MARL) is introduced to improve the scalability of RL agents by using a decentralized control framework. \citet{chu2019multi} use advantage actor-critic (A2C) as a large-scale TSC method. To be specific, neighbors' information is adapted to improve sample efficiency and promote cooperative strategy. Furthermore, a spatial discount factor is introduced to improve the learning efficiency, i.e. to reduce fitting difficulty. To enable cooperation of traffic signals, recent works study how to encourage cooperation through graph representation learning. \citet{wei2019colight} propose to use a graph attention neural network in the setting of large-scale road networks with hundreds of traffic signals. They model each TSC as an agent. Agents learn to communicate by attending to the representations of neighboring intersections. Their results demonstrate the effectiveness of the attention mechanism to help cooperation and achieve superior performance over state-of-the-art methods. Concurrently, \citet{devailly2021ig} further exploit the vehicular data at its finest granularity by representing every vehicle as a node. They demonstrate the flexibility of GCNs, which can enable transferability to unseen road networks. However, neither of these works evaluates their methods under exogenous uncertainties.

\subsection{Robustness in Traffic Signal Control} There are several factors that could affect the model's robustness, such as sensor failures and demand surges. 
In transportation research, a very straightforward way to solve the exogenous uncertainty problem from sensor failure is to use imputation methods~\citep{tang2015hybrid,chen2019missing,chen2021low}. 
For example, recent work uses 
a variational Bayes approach to predict missing values accurately~\citep{chen2019missing}. 
Graph Neural Network (GNN) can also be an efficient and effective tool for recovering information from malfunctioning sensors \citep{wu2020inductive}.
Bayesian multiple imputation and bootstrap have also been used to approximate the distribution of the training set in order to estimate the state-action value function given missing data \citep{lizotte2008missing}. 

Such methods are tailored to sensor failures and do not solve problems related to demand surges and different road networks. Therefore, we do not focus on imputation methods here.

Recently, deep RL has proved to be robust in small-scale networks under the impact of special events, such as demand surges, sensor failures, and partial detection. \citet{rodrigues2019towards} developed the callback-based framework to enable flexible evaluation of different deep RL configurations under special events. They concluded that when training in scenarios with sensor failures, the RL approach can be quite robust to the wide sensor failure and demand surge problems. \citet{zhang2020using} demonstrate that deep RL agents can be robust within the partially detected intelligent transportation systems (PDITS), which is a partially observable Markov decision process (POMDP) in the RL community, 
in which only part of vehicle information can be acquired. They have conducted experiments under different detection rates and report that the RL-based control method can improve travel efficiency even with a low detection rate. However, their evaluation scenario is limited to one to five intersection cases. Most importantly, they have not further discussed how to improve the robustness based on previous reinforcement learning methods. Our model can be extended to a large-scale network.  \citet{ghanadbashi2023handling} introduces a model called OnCertain to improve decision-making in self-adaptive systems that interact with each other in dynamic environments. The proposed system can handle uncertainty caused by unpredictable and rare events while having limited information about the environment.

\subsection{Generalization in Traffic Signal Control}


The training mechanism for Deep RL follows a trial-and-error approach and is computationally expensive~(see chapter 4 in \cite{sutton2018reinforcement}). For traffic signal control, training models on large-scale networks or using a variety of different traffic demands quickly becomes prohibitive~\citep{wei2019colight}.
As a result, designing methods that can learn on smaller networks and transfer their knowledge to large-scale ones can be beneficial. 

Recently, meta-RL\footnote{meta-RL: a learning-to-learn approach that involves learning on training tasks in order to ease training on test tasks drawn from the same family of problems.} has been applied to traffic signal control problems. \citet{zang2020metalight} propose to use value-based meta-reinforcement learning for traffic signal control which includes periodically alternating individual-level adaptation and global-level adaptation. Based on the previous work~\citep{zang2020metalight}, \citet{zhu2023metavim} take the policies of neighbor agents into consideration and consider learning a latent variable to represent task-specific information to not only balance exploration and exploitation but also help learn the shared structures of reward and transition across tasks. \citet{zhang2020generalight} design a WGAN-based~\citep{arjovsky2017wasserstein} flow generator to generate different traffic flows to improve the generalization ability of TSC models to different traffic flow environments. However, MetaLight~\citep{zang2020metalight} considers training on larger-scale networks, then testing on a subset of training networks or smaller networks. Recently, GNNs have demonstrated generalizability to different road structures and traffic flow rates or demands.~\citet{nishi2018traffic} stack multiple GCN layers onto neural networks to improve the generalizability to different vehicle generation rates during training. \citet{wei2019colight} use graph attentional networks to facilitate communication and promote cooperation among intersections. \citet{devailly2021ig} represent traffic entities as nodes in the graph to enable generalizability  to new road networks, traffic distributions, and traffic regimes.

\subsection{Summary of Previous Work on Robustness and Generalizability for Traffic Signal Control}

Table~\ref{survey} summarizes and compares the previous works with respect to the following aspects: 1. Generalizability to different networks and traffic flows or demands, and 2. Robustness to sensor failures (noise).

Deep reinforcement learning methods have demonstrated robustness to sensor failures~\citep{tan2020robust,rodrigues2019towards}. Furthermore, by using the transfer learning technique~\citep{tan2020robust}, the trained model can also handle demand surges. However, the above methods do not adapt to new road networks. At best these methods require a fine-tuning step before being deployed on a new network.

Some work proposes using meta-learning to improve the generalizability to different road networks and traffic flow distributions~\citep{zang2020metalight,zhu2023metavim,zhang2020generalight}. However, the training data sets usually include more scenarios than the testing sets, or the testing sets are a subset of training sets~\citep{zang2020metalight}. Furthermore, MetaLight~\citep{zang2020metalight} still needs to re-train its model parameter on new intersections. As a result, they cannot perform zero-shot transfer to new road networks.

Recently, graph-convolutional networks have demonstrated their ability to further improve generalizability, enabling zero-shot transfer learning to new road structures and traffic settings that have never been experienced during training. In summary, IGRL~\cite{devailly2021ig} is the only work that can enable zero-shot transfer learning for new scenarios. Therefore, we choose the IGRL model and its variant as our reinforcement learning baseline methods.

In this work, we build upon the previous work~\citep{devailly2021ig} and systematically evaluate the transferability of IGRL. We are the first to jointly improve generalizability to different networks and robustness to sensor failures and demand surges.


\begin{table*}[ht]
\tiny
\caption{Previous works address generalization and robustness separately. RGLight, the method proposed in this paper, studies their combination.}\label{survey}
\centering
\begin{tabular}{c}
\begin{tabular}{ l c c c}
\toprule
	 Method & Disjoint train \& test  & Varying Traffic  & Sensor failure \\ %
     &  networks &  flows (demand) & (noise) \\
        \midrule
	   MetaLight~\citep{zang2020metalight}  & \halfcirc &  \halfcirc  &\emptycirc \\ 
	   MetaVIM~\citep{zhu2023metavim}& \halfcirc &  \halfcirc & \emptycirc \\
	   GeneraLight~\citep{zhang2020generalight} &\halfcirc&  \halfcirc& \emptycirc \\
	  GCN + RL~\citep{nishi2018traffic}&  \emptycirc &   \emptycirc  &\emptycirc \\ 
	  CoLight~\citep{wei2019colight} & $\bigcirc$ &  \emptycirc & \emptycirc \\ 
	  IGRL~\citep{devailly2021ig}& \fullcirc &  \fullcirc &\emptycirc \\ 
	  Transfer learning+Dueling DQN~\citep{wu2020multi}&  \emptycirc&  \fullcirc& \emptycirc \\ 
	   Call-back based Deep RL~\citep{rodrigues2019towards}& \emptycirc& \emptycirc&\fullcirc \\ 	
	   Robust TSC~\citep{tan2020robust} & \emptycirc &  \emptycirc&\fullcirc \\ 
	   Interpolation-based robust feedback controller~\citep{komarovsky2019robust} & \emptycirc &  \fullcirc &\emptycirc \\ 
    \midrule
    RGLight (this paper) & \fullcirc &  \fullcirc &\fullcirc \\

	\noalign{\smallskip}\bottomrule
\end{tabular}
\end{tabular}
    \begin{tablenotes}   
         \item[] \fullcirc~: investigated; \emptycirc ~: not investigated; \halfcirc~: partly investigated. In particular, Meta-learning methods generalize to different networks or different traffic flows by re-training the model parameters given the new network. In other words, they do not perform zero-shot transfer learning.
    \end{tablenotes}
\end{table*}

\section{Methodology}



\begin{figure}[h]
\center
\includegraphics[width=8.5cm]{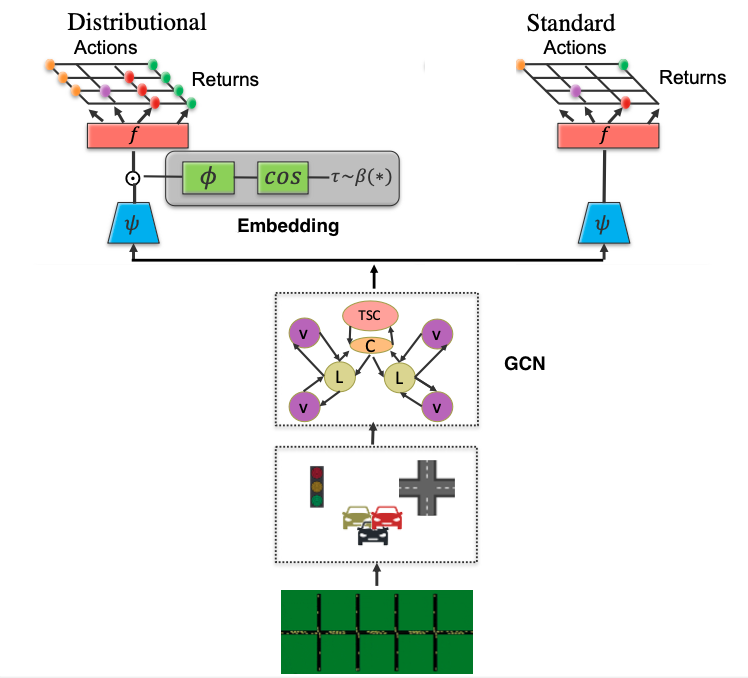}
\caption{Framework overview (inspired by~\citet{dabney2018implicit}). The graph (nodes and edges) encodes the structure of the road network. The current state of the road network at each time step is encoded as node features in this graph. The graph is modeled using a graphical convolutional network (GCN). The parameters of the GCN are learned using one of two objectives. Either the standard RL objective~\citep{devailly2021ig} which estimates pointwise state-action returns. Either the distributional RL objective for which multiple samples (left branch, multiple points per action/color) are drawn from quantiles and implicitly define the distribution of state-action returns for all actions (right branch, one point per action/color). In both cases, an embedding function $\psi$ is used followed by a non-linear layer (not represented on the figure) to provide the value function $Q(s,a)$. In the distributional RL case, the embedding is combined with a quantile embedding $\phi$. Mathematical details are provided in Sections~\ref{sec:gcn} and~\ref{dismarl}.}
\label{dgrl}
\end{figure}

The proposed framework is shown in \Cref{dgrl}. Like \citet{devailly2021ig}, we first encode the road network around each TSC including the moving components as a graph with nodes and edges. We abstract each vehicle feature (V), lane feature (L), connection feature (C), and traffic signal controller (TSC) feature as nodes of the  graph (\Cref{agentdesign}). Then a representation of the graph is learned using a graph convolutional network (GCN), see \Cref{sec:gcn}. 

We train the GCN to estimate state-action values (or returns) either using a standard RL objective (\Cref{sec:gcn}) or a DisRL objective (\Cref{dismarl}). In standard RL, the GCN provides a graph representation embedding $\psi$ (\Cref{dgrl} right branch). In DisRL, we combine the embedding with an \emph{embedding function} $\phi(\cdot)$ (\Cref{dgrl} left branch). We then combine the values of the returns estimated by the DisRL and the standard RL objectives (\Cref{robustloss}). 

The combined estimated returns can then be decoded (greedily) to obtain the agent's action. Once an action $a_t$ is executed, the environment changes (e.g., following a micro-traffic simulator) and the agent can then pick its next action ($a_{t+1}$). In practice, we assume that the agent can execute an action every second (i.e., a timestep lasts one second). 


From Figure~\ref{dgrl}, we can find  that on the right (traditional DQN/IGRL), pointwise estimates of state-action returns are used (one point per action/color) while on the left, multiple samples (i.e. multiple points per action/color) are drawn from quantiles and implicitly define the distribution. of state-action returns for all actions.

\subsection{Agent Design}
\label{agentdesign}
\subsubsection{State space}
Given the state observation for each signal controller $i$, the state-action pairs for each TSC are denoted
$(s_i, a_i) \in S \times A, i= 1,\ldots ,K$. 

We assume that there are $K$ intersections in the system and each agent, i.e., TSC, can observe part of the system state $s \in S$. The number of layers in the GCN defines how large the observable part of the state space is for a given agent. For instance, when using only 2-3 layers, given the architecture of the GCN, only information regarding a local intersection (connectivity features corresponding to controllable connections and traffic features corresponding to immediately inbound and outbound lanes) is perceivable to that intersection's agent.
Based on~\citep{devailly2021ig}, we consider the following features in each entity:

\begin{itemize}

\item {\textit{TSC feature:}} represents the state of a controller. The features are the number of seconds since a traffic controller performed its last phase switch.

\item{\textit{Connection feature:}} represents the state of an existing link between an entry lane and an exit lane. For example, the connection exists between an entry lane A and an exit lane B if a vehicle on lane A is allowed to continue its travel to lane B. The features in the connection feature are whether a connection is opened under the current phase; whether an open connection between an entry and an exit lane has priority or not; the number of switches the controller has to perform before the next opening of a given connection; and whether the next opening of the connection will have priority or not.

\item {\textit{Lane feature:}} represents the state of a lane. It includes the length of the lane.
\item {\textit{Vehicle feature:}} represents the state of a vehicle which includes its current speed and position on the current lane as a feature.
\end{itemize} 
\subsubsection{Action space}  At every intersection of the road network, there is a predefined logical program, composed of a given number of phases, depending on the roads, lanes, and the connection information. The program is given by the road network. The binary action of the agent is either to switch to the next phase or prolong the current phase. This  modelling is compatible with TSCs using different programs.

\subsubsection{Reward function} Each agent $i$ obtains a reward $r^t_i$ at time $t$ from the environment. In this paper, we want to minimize the travel time of the vehicles. The reward is defined as the negative sum of total queue lengths per intersection $q$, $r^t_i=-\sum_{l} q^t_{i,l}$. where     
$ q^t_{i,l}$ is the queue length on the lane $l$ at time  $t$.


\subsection{Graph Representation Learning on Different Nodes}\label{sec:gcn}

\subsubsection{Graph representation using a GCN}
As in \citet{devailly2021ig}, we encode the state of the network as a graph. Traffic signal controllers, lanes, connections between lanes, and vehicles are nodes in this graph. Edges connect nodes that are adjacent on the road network (e.g., a vehicle node to its current lane node or a lane node to its connections with a neighbor lane). 

The graph is encoded using its adjacency matrix $A$ and it is processed by a graph convolutional network (GCN)~\citep{Kipf2016,liu2020introduction}. The GCN propagates information between nodes to obtain a representation $H^n$ at each layer $n$:
\begin{equation}
H^{n+1}=\sigma\left(D^{-\frac{1}{2}} A D^{-\frac{1}{2}} H^{n} W^{n}\right),
\label{gcn}
\end{equation}
where $D$ is a (diagonal) degree matrix ($D_{ii}=\sum_j A_{ij})$ which normalizes $A$ using its number of neighbors, $W^{n}$ are learned parameters and $\sigma$ is the sigmoid activation function~\citep{Kipf2016}. 

Along with the graph structure, nodes and edges can have features $X$.
These features are used to obtain the first-layer representation:
\begin{equation}
H^0 = \sigma(W^{0\top}X+b^{0})
\end{equation}
where $W^0$ and $b^0$ are learned parameters. 

Assuming $N$ hidden layers, we use the last-layer representation $H^N$ to predict a value function. Let $\psi: 
\mathcal{X} \rightarrow 
\mathbb{R}^{d} $ 
be an embedding function parameterized by the GCN layers. We add a subsequent fully-connected layer to map $\psi(x)$ to the estimated action values, such that $Q(x,a) \equiv f(\psi(x))_a$, where $a$ in $f(\cdot)_a$ indexes the output action.
We can get the estimated Q values as:
\begin{equation}
Q(s,a)= (H^{N} W_{p}+b_{p})_{(s,a)},
\label{qvalues}
\end{equation}
where $W_p \in R^{c \times p}$ and $b_p \in R^{p}$ are parameters of the neural networks, and $p$ is the number of phases (action space).


In Deep RL, the objective to optimize at each time step $t$ is 
\begin{equation}
\mathcal{L}( \theta)= \big(y_t-Q \left(s_{t}, a_{t};\theta  \right)\big)^2,
\label{smoothl1loss}
\end{equation}
where  $y_t=r_t +\gamma max_{a} Q(s_{t+1}, a_{t+1})$,  $\theta$ represents all trainable parameters ($b^0, W^{0,\ldots ,N-1}, b_p, W_p$) and $\gamma$ is the (fixed) discount factor.

The (greedy) action associated with the value function can be obtained for each state as:
\begin{equation}
\pi(s)=\underset{a \in \mathcal{A}}{\arg \max }~Q(s, a).
\label{optimalact}
\end{equation}
where $\pi(s)$ denotes the policy in state $s$.

\subsubsection{Parameter sharing}
Each TSC learns to maximize its local reward and as such TSCs are independent. However, the parameters of all TSCs are shared  to encourage learning parameters that transfer to a variety of situations. In particular, nodes of the same type both within the same TSC and across TSCs share the same parameters. Parameter sharing also reduces the memory footprint of the system (since the number of parameters is now independent of the number of TSCs). The system can then scale to very large networks \citep{devailly2021ig}.  

\subsection{Distributional RL}\label{dismarl}

The previous section introduces standard RL for GCNs (\ref{smoothl1loss}). Now, we discuss learning the GCN model using distributional RL (DisRL). Compared to traditional RL, DisRL models the distribution over returns. The expectation of that distribution yields the standard value function. In this work, we use implicit quantile networks~\citep{dabney2018implicit}, a distributional version of Deep Q-Networks~\citep{silver2017mastering}. Implicit quantile networks can approximate any distribution over returns and show superior performance compared to other DisRL methods~\citep{bellemare2017distributional,dabney2018distributional}.



Implicit quantile networks define an implicit distribution using samples $\tau$ from a base distribution $\tau \sim U([0,1])$). The implicit distribution is parameterized using $\phi:[0,1] \rightarrow R^d$. The function $\phi$ provides the embedding for quantile $\tau$. This embedding  $\phi$ is combined with the GCN's output embedding $\psi$ to form the approximation of the distributional Q-values (see Figure~\ref{dgrl} (a)):
\begin{equation}
Z_{\tau}(s, a) \equiv f(\psi(s) \odot \phi(\tau))_{a},
\label{iqn}
\end{equation}
where $\odot$ represents the element wise product, the $a$ on the RHS indexes the output of the function $f$. We use the same embedding function as in~\citep{dabney2018implicit}:
\begin{equation}
\phi_{j}(\tau):=\operatorname{ReLU}\left(\sum_{i=0}^{n-1} \cos (\pi i \tau) w_{i j}+b_{j}\right),
\label{embeddingfunc}
\end{equation}
where $n$ is the size of the input embedding, $j\in 1,\ldots,d$ indexes different units (neurons), and $w_{ij}$ and $b_j$ are parameters shared across all TSCs (much like parameters of the GCN \Cref{gcn} are also shared across TSCs).

As a result, the state-action value function can be represented as the expectation:
\begin{equation}
Q(s, a):=\underset{\tau \sim U([0,1])}{\mathbb{E}}\left[Z_{(\tau)}(s, a)\right], 
\label{zexpectation}
\end{equation}
and its associated greedy policy can be obtained from \Cref{optimalact}.

In DisRL, we want to minimize the distance between two distributions so as to minimize the temporal difference  error (TD-Error). For two samples $\tau , \tau' \sim U([0,1])$, and policy $\pi$, the TD-Error at time step $t$ can be computed as:
\begin{equation}
\delta_{t}^{\tau, \tau^{\prime}}=r_{t}+\gamma Z_{\tau^{\prime}}\left(s_{t+1}, \pi\left(s_{t+1}\right)\right)-Z_{\tau}\left(s_{t}, a_{t}\right).
\label{tderror}
\end{equation}

Furthermore, the random return is approximated by a uniform mixture of $K$ Dirac delta function:
\begin{equation}
Z_{}(s,a):=\frac{1}{K}\sum_{i=1}^K \delta_{\mu_i (s,a)},
\label{ndirac}
\end{equation}
where each $\mu_i$ assigned a fixed quantile target. The quantile target's estimations are trained using the Huber loss~\citep{crow1967robust} with threshold $\lambda$.

As a result, the distributional version of loss function is formulated as:
\begin{equation}
\mathcal{L}_{dis}\left( \theta \right)=\frac{1}{M^{\prime}} \sum_{i=1}^{M} \sum_{j=1}^{M^{\prime}} \rho_{\tau_{i}}^{\lambda}\left(\delta_{t}^{\tau_{i}, \tau_{j}^{\prime}}\right),
\label{iqnloss}
\end{equation}
with $\rho_{\tau_{i}}^{\lambda}$ is the quantile regression term~\citep{dabney2018implicit}, $M$ and $M'$ the number of samples used to evaluate the TD-error. 



\subsection{RGLight}
\label{robustloss}

In the previous sections, we introduce two different reinforcement learning formulations for learning TSC policies (see \Cref{dgrl}).  
Our initial experiments
show important empirical differences between the two approaches. 

First, we find that distributional RL converges faster than classical RL  in our domain. We also note that the embeddings learned by both approaches are different (see Figure~6 in the supplementary material for an example). 

We suspect a combination of the learned policy might yield the best of both worlds.
To do so, we train both approaches separately and then combine their (estimated) Q-values (during testing) (see Figure~\ref{qvalues}).  

Given a set of actions $A(s_t)=\{a[1],...,a[n]\} $, The estimated Q-value for action $a_i$ is $Q(s_t,a_i)$ at time $t$. 
We first normalize the Q values of both methods. We find that exponentiating the values first yields better results~\citep{wiering2008ensemble}:
\begin{equation}
\tilde{Q}(s,a)= \frac{e^{Q(s,a)/T}}{\sum_i e^{Q(s,a_i)/T}}.
\label{combineloss}
\end{equation}
%
We then obtain $\tilde{Q}^{RG}$ the Q-value used by RGLight as a convex combination of the normalized Q-values of the two methods:
\begin{equation}
\tilde{Q}^{RG}=\kappa \tilde{Q}^{deter}+(1-\kappa)\tilde{Q}^{dis},
\label{combineloss}
\end{equation}
where we dropped the $s$ and $a$ indexes for clarity and $\kappa\in[0,1]$ is the relative importance of the standard RL approach. 
We ensemble the prediction results from two frameworks to improve the robustness and generalizability of our model. Based on preliminary simulations, we find that $\kappa=0.6$ and $T=5$ offer more consistent and higher performance across experiments. 


\section{Experiments}

In this section, we study the effectiveness of the  RGLight method for multi-agent TSC. We aim at answering the following questions:
\begin{itemize}
    \item How does the proposed method perform compared with other state-of-the-art baselines? (\Cref{differtregime} and \Cref{sensorfail})
    \item Is the proposed method more robust to sensor failure problems compared to other baseline methods? (\Cref{differtregime} and \Cref{sensorfail})
    \item Can the proposed method generalize to different road network structures and traffic regimes? (\Cref{generalization ability})
    \item How can we balance the trade-off between representation capacity and learning stability to improve the overall robustness and generalizability? (\Cref{generalization ability} and \Cref{sensorfail})
\end{itemize}

\subsection{Experiment Setup}
\label{setupexp}
The scenario we study is one where a  system learns in a ``controlled environment'' on synthetic networks with no missing data. Then the performance, robustness, and generalizability of the system are tested by ``deploying'' it in a more realistic scenario that involves new networks (synthetic or from the real world), different traffic regimes (demand surges), and missing data. A visualization of the learning setup is shown in Figure~\ref{network}. 

To be more precise, we train RL methods (DGRL, IGRL, and GNN-TSC) on synthetic road networks for 60 episodes without missing data or demand surge. Then we test their performance on either other synthetic networks or, perform zero-shot generalization by controlling the TSCs of two real-world networks (a part of Luxembourg and Manhattan). All of our studies use the simulation of urban mobility (SUMO) \citep{krajzewicz2002sumo} micro simulator.


\subsubsection{Background and Assumption}
\label{backgroundandassump}
\begin{itemize}

    \item \textbf{Sensor Failures:} In all of our experiments, we assume that \textit{we know the lane each vehicle is in}. We imagine, for example, that on each traffic signal controller, there would be a camera/detector that can sense which vehicle has entered which lane, and it is not likely to fail~\citep{wu2020multi}.
    The most common cause of missing data comes from the sensor failure of probed vehicles, which means that the system detects the vehicle, but does not get its current speed and exact position \citep{lu2008faulty,qiu2010estimation}. We assume faulty vehicle sensors provide a value of zero. 
    
    \item \textbf{Traffic flows:} We consider different traffic flows as both different traffic distributions and traffic  demands. Particularly, different traffic demands are based on the arrival rate. For all these experiments, the trip is generated by SUMO's trip generator.\footnote{https://sumo.dlr.de/docs/Tools/Trip.html} The arrival rate is controlled by the option \textit{period} in SUMO \citep{krajzewicz2002sumo}. By default, this generates vehicles with a constant period and arrival rate of (1/period) per second. Note that for different scales of road networks, the same arrival rate will end up with different traffic signal performances.\footnote{To obtain a fair comparison, we consider the heavy traffic regime as two times the normal traffic regime in simulated data. In our experiment, we set the normal traffic regime with period=4 and the heavy traffic regime with period=2.}  For the trip distribution, the number of departures per second will be drawn from a binomial distribution. In our experiment setting, the trip distribution (the probability of a successful departure) will be changed every 120 seconds. As a result, both the traffic distribution and the traffic demands can be changed in our study.


    \item \textbf{Evaluation metrics:} We discuss the performance of the methods using several standard evaluation metrics (~\cite{devailly2021ig,wei2018intellilight}).
    
    \subsubsection*{Travel time} The travel time is defined as the time duration between the real departure time and the time the vehicle has arrived. The information is generated for each vehicle as soon as the vehicle arrives at its destination and is removed from the network.
    \subsubsection*{Queue length} The queue length is calculated at the lane level using the end of the last standing vehicle. This criterion measures congestion, representing whether it significantly slowed close to an intersection.
    \subsubsection*{Delay} The delay $d_t$ measures the gap between the current speed of the vehicle and its maximum theoretically reachable speed, which is constrained by the type of the vehicle and the maximum allowed speed on the current lane
    \begin{equation}
    s_{v}^{*}=\min \left(s_{v^{*}}, s_{l}\right),
    \label{delay1}
    \end{equation}
    \begin{equation}
    d_{t}=\sum_{v \in V}\left(s_{v}^{*}-s_{vt}\right) / s_{v}^{*}
    \label{delay2}
    \end{equation}
    where $V$ is the total number of vehicles traveling in the current network, $s_{v^*}$ is the maximum speed that the vehicle can reach, $s_l$ is the speed limit of this road, and $s_{vt}$ is the vehicle speed at time step $t$ and $d_t$ denotes the delay at time $t$.  Instantaneous delay for 1 vehicle is how far it currently is from its optimal theoretically reachable speed 

\end{itemize}

\subsubsection{Datasets}
We evaluate the different methods using both synthetic networks with synthetic data and real-world networks with real-world traffic routes.

\begin{figure}[h]
\center
\includegraphics[width=9cm]{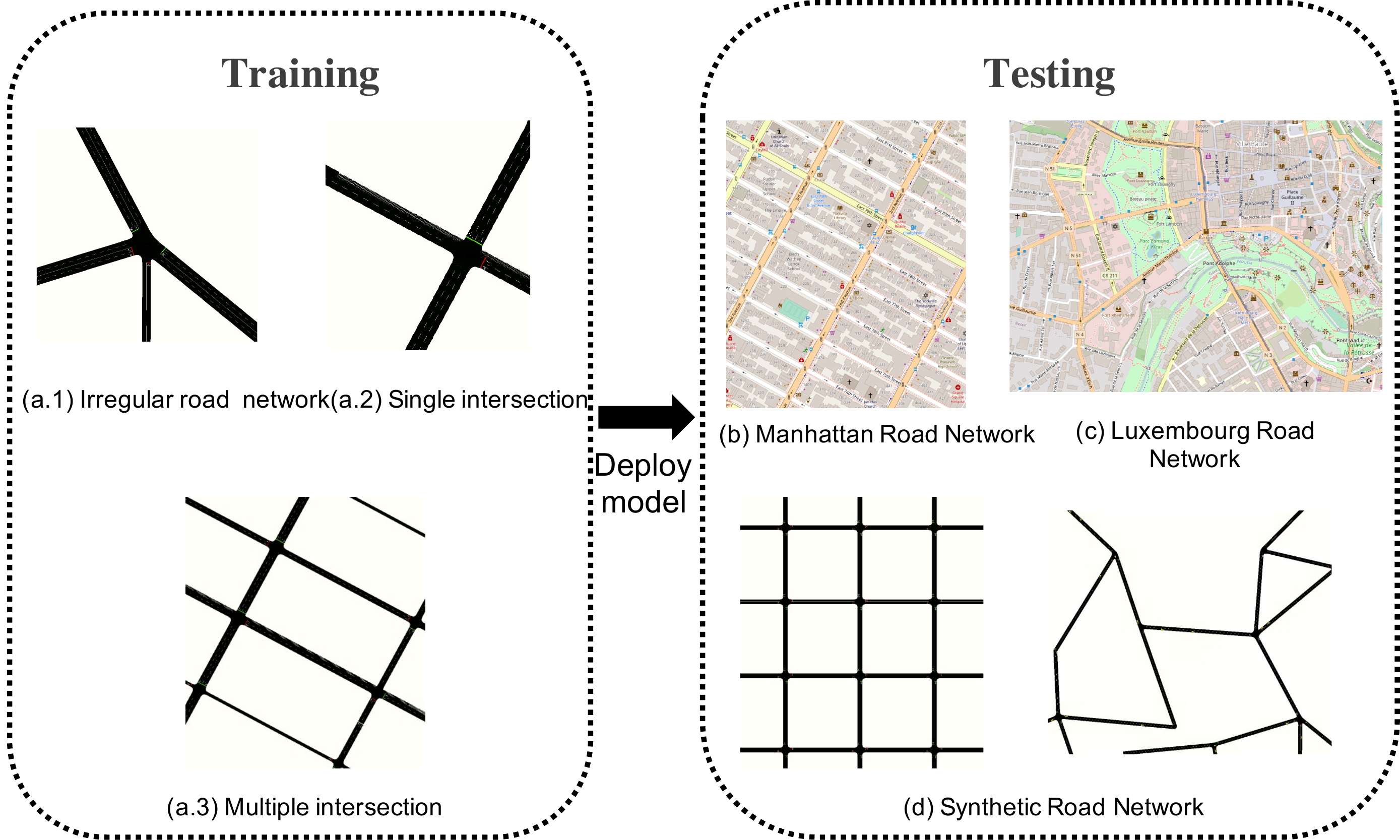}
\caption{Learning scheme for our model. Diverse synthetic road networks are used for the training set while real-world road networks are used for the testing set.}
\label{network}
\end{figure}

\begin{itemize}
    \item Synthetic networks:
    We use the same approach to generate the synthetic networks as in IGRL~\citep{devailly2021ig}. The structure of the synthetic road networks is generated at random using the SUMO simulator, the number of intersections varies between two and ten; the length of every edge is between 100 and 300 meters, and the number of lanes per route is between one and four. Some examples of the generated networks can be seen in Figure~\ref{network}. We try to maximize the variability of the training networks by generating random networks to cover the most typical cases in real-world networks.
    \item Real-world networks:
    We use representative traffic data\footnote{Luxembourg: \url{https://github.com/lcodeca/LuSTScenario}, Manhattan: \url{https://traffic-signal-control.github.io/}} from part of Luxembourg and Manhattan to evaluate the performance of our model in real-world settings. Manhattan  has a grid-like road network and contains 75 traffic lights and 550 intersections. The Luxembourg network  contains 22 traffic lights and 482 intersections. It is also more irregular than Manhattan.  Both networks have different traffic demand evolution characteristics as shown in Figure~1 and~2 in the supplementary material.
\end{itemize}

\subsubsection{Baselines}
We compare our method with several state-of-the-art methods, including both classical transportation methods and learned ones. 

\textbf{Transportation Methods}:
\begin{itemize}
    \item Fixed time Baseline~\citep{koonce2008traffic}: It uses a predetermined plan for cycle length and phase time. This technique is widely used when the traffic flow is steady~\citep{koonce2008traffic}.
    \item Max-moving-car-dynamic-heuristic (Greedy): This dynamic heuristic-based method aims at ensuring that as many vehicles as possible are moving on inbound lanes at any given time, in the spirit of the popular baseline \emph{Greedy} \citep{varaiya2013max} under a cyclic setting. Controllers switch to the next phase if, on inbound lanes, the number of stopped vehicles is superior to the number of moving vehicles, and prolongs the current phase otherwise.
\end{itemize}   

\textbf{Reinforcement Learning Methods}:
\begin{itemize}
    \item Inductive Graph Reinforcement Learning (IGRL) \citep{devailly2021ig}: This recent approach uses graph convolutional networks with a decentralized RL objective. The authors show that their approach can scale and transfer to massive-scale networks. Our robust learning framework is based on IGRL. We compare against their best-performing model IGRL-V which models vehicles as nodes.
    \item Graph Neural Networks for TSC (GNN-TSC) \citep{wei2019colight}: Similar to IGRL, the authors propose a GNN-based RL-trained model. Compared to IGRL~\citep{devailly2021ig}, the method does not consider individual vehicles as nodes in the graph. Instead, they model information at the lane level. 
    With that in mind, we use IGRL-L, a version of IGRL that models lane nodes rather than vehicles as nodes. This version is similar to the CoLight method~\citep{wei2019colight}.\footnote{The authors of \citep{wei2019colight} rely on the CityFlow simulator \url{https://cityflow-project.github.io/}, we use SUMO, which prevents a direct comparison without a major code rewrite.} 
    \item Independent Reinforcement Learning (IRL): An independent deep Q-Learning (DQN) agent can be used to model each TSC. DQNs have som level of robustness given demand surges and sensor failures~\citep{rodrigues2019towards,zhang2020using}. Further, the IRL baseline couples DQNs with recent developments for improved robustness: double Q-Learning \citep{hasselt2010double}, a dueling architecture~\citep{wang2016dueling}, and noisy layers~\citep{fortunato2017noisy}.
    
\end{itemize}

\subsection{Performance Comparison}
In this section, we compare the performance of the above baselines to the performance of RGLight with respect to different traffic regimes and sensor failures. All experiments are repeated 30 times with different random seeds for trip generations and the average results are presented. For every evaluation metric, we report the sum of a 1,000-time-step simulation. 
Note that for each criterion, for readability, the obtained value is divided by 100 in the tables. We also provide a video illustrating the different methods.\footnote{Simulation video link: \url{https://youtu.be/wTUkoXvVghs}}

\begin{table*}[ht]
\tiny
\caption{Comparison result under different traffic regimes (average and standard deviation in seconds). In this experiment, we use synthetic traffic data to better control the traffic demand surge, where the heavy regime's traffic demand is twice the normal traffic regime. Lower is better, and the best mean value is bolded.}\label{traffic regim}
\begin{center}
\begin{tabular}{c}

\toprule
Normal regime \qquad\qquad\qquad\qquad\qquad \qquad\qquad \qquad\qquad Heavy regime \\
\begin{tabular}{cccc|ccc}
\midrule \noalign{\smallskip}
	Methods  & Delay & Queue length & Travel time & Delay & Queue length & Travel time \\ %
	\noalign{\smallskip}\hline\noalign{\smallskip}
	Fixed time   & 789.26(36.36)   &588.88(35.39) &   1182.26(125.57) & 4059.19(108.54) & 4553.34(112.34) & 13901.72(922.15)\\
	\emph{Greedy}   & 379.91(12.22)  &191.91(10.41) &  670.28(32.55) & 6201.11(183.23) & 6865.94(190.42) & 15150.86(734.36)\\
	\noalign{\smallskip}\hline\noalign{\smallskip}

	IRL   &  1257.58(31.84)   &1013.89(29.40) & 1242.38(46.78)  & 5257.58(152.62) & 6670.75(160.25) & 14112.98(498.12)\\
	GNN-TSC  &311.85(4.32) &210.43(10.53) &517.15(34.32) &  2998.63(61.47)  &3645.75(92.68)  &6092.63(428.75)   \\
	IGRL  & 288.16(8.66)   &125.89(7.72) & \textbf{501.36(22.22)}  & 2962.92(81.81) & 3515.23(86.00) & 6051.32(355.51)\\
    RGLight & \textbf{244.15(4.25)}  & \textbf{80.11(2.74)} & 501.95(20.77)   & \textbf{2503.96(71.91)} & \textbf{3029.45(76.57)} & \textbf{5030.31(313.82)}\\
	\noalign{\smallskip}\bottomrule

\end{tabular}

\end{tabular}

\end{center}
\end{table*}

\subsubsection{Comparison under Different Traffic Regime in Synthetic Networks}
\label{differtregime}

Table~\ref{traffic regim} reports the performance of different methods for both normal and heavy traffic regimes in synthetic networks.\footnote{We conduct the demand surge experiment in a synthetic network because it is difficult to control the demand parameter in real networks with real traffic demand.} We use the same road network (not seen in the training set) in tests for all methods with 30 random seeds for trips.

Overall, RGLight outperforms others in the normal regime across the three metrics except in terms of travel time where IGRL does as well. RGLight also shines in a heavy regime showing that it is more robust to demand surges.

We see that Fixed time does not perform as well as \emph{Greedy} in normal traffic regimes but better than \emph{Greedy} in heavy traffic regimes. In terms of travel time, RGLight performs about the same as IGRL in the normal regime. 
As shown in Figure~\ref{tripdur}, although IGRL and RGLight provide similar average travel times, the empirical distribution of their difference is skewed to the right. This seems to indicate that under this evaluation RGLight is more equitable. In a heavy traffic regime, we see that RGLight outperforms IGRL by a large margin. 

\begin{table*}[ht]
\centering
\tiny
\caption{Comparison result under missing values in Manhattan road network (average and standard deviation in seconds). These two experiments with real-world road networks can test not only test the robustness of different methods but also test how they generalize to different road networks since we train our model on smaller synthetic networks. }\label{table:mh}
\begin{center}
\begin{adjustbox}{max width=\textwidth}
\begin{tabular}[]{ c c c c}
\toprule
\multirow{2}*{Methods} &\multicolumn{3}{c}{Missing Probability (20/ 40/ 60 \%)}  \\
\cline{2-4}
~ & Delay & Queue Length & Travel time \\ 
\midrule
Fixed time & 1356.45(41.29) & 937.47(40.48)  &  1871.86 (238.99)\\
 
\emph{Greedy} & 1144.30(34.32)  &  907.24(44.43)& 1630.67(264.48) \\
\midrule

GNN-TSC & 484.49(4.84) / 497.18(9.61) / 696.15(9.82)&469.75(7.84) / 578.98(7.68) / 612.96(5.24) &973.46(27.23) / 1273.31(12.67) / 1346.75(41.45)\\

IGRL & 413.94(9.94) / 518.41(11.87) / 653.22(13.76) & 314.74(3.96)
/ 417.93(3.36) / 499.89(3.55)
  & 966.65(25.47) / 1163.89(10.32) / 1260.46(18.27) \\

RGLight & \textbf{364.23(3.95)} / \textbf{397.91(4.05)} / \textbf{492.89(9.12)} &\textbf{311.99(3.01)} / \textbf{363.60(3.17) } / \textbf{403.11(3.22)} & \textbf{954.28(15.66)}/ \textbf{1032.58(13.63)} / \textbf{1088.67(17.36)}\\

\bottomrule

\end{tabular}
\end{adjustbox}
\end{center}
\end{table*}

\subsubsection{Comparison under Sensor Failures in Different Real-world Road Networks}
\label{sensorfail}

\begin{table*}[ht]
\tiny
\centering
\caption{Comparison Result under missing values in Luxembourg  road Network.}\label{table:lx}
\begin{center}
\begin{adjustbox}{max width=\textwidth}
\begin{tabular}[]{ c c c c}
\toprule
\multirow{2}*{Methods} &\multicolumn{3}{c}{Missing Probability (20/ 40/ 60 \%)}  \\
\cline{2-4}
~ & Delay & Queue Length & Travel time \\ 
\midrule
Fixed time & 594.22(16.24)  & 509.79(14.33)  &  620.98(68.54)\\
 
\emph{Greedy} & 754.27(22.16)  &661.03(19.97)&781.38(131.84) \\
\midrule

GNN-TSC & 489.50 (6.38) / 595.84(8.82) / 723.65(10.79) & 385.65 (5.06) 
/ 511.68 (8.71) / 627.66(10.59) & 534.16(29.69) / 651.36(49.48) / 721.98(58.02)\\

IGRL &  438.26 (8.31) / 531.25(9.30) / 678.75(14.37) & 373.33 (4.89)
/ 460.07 (6.23) / 589.61(7.35) & 527.38(31.20) / 591.92(32.71) / 683.25(40.51) \\

RGLight & \textbf{419.43(6.23)} / \textbf{501.86(7.12)} / \textbf{545.68(8.56)} & \textbf{356.28(3.27)} / \textbf{421.85(5.71)} / \textbf{469.28(7.91)} & \textbf{467.94(16.35)} / \textbf{535.66(23.98)} / \textbf{572.67(28.01)}\\

\bottomrule
\end{tabular}
\end{adjustbox}
\end{center}
\end{table*}

In this experiment, we test our model's performance with two real-world road networks using real traffic demand (see Figure~1 and ~2 in supplementary material). The IRL method does not scale to such large networks (the parameters increase linearly with the number of TSCs) and so we cannot report its performance. Transportation baselines do not consider speed or vehicle position and so their performance is robust to noisy sensors.

We first discuss the performance in the Manhattan road network from \cref{table:mh}.
We find RGLight outperforms other methods. It is also more robust in scenarios with higher proportions of missing data compared to the other RL baselines.  


Second, we study methods on Luxembourg's road network. 
Results in \cref{table:lx} are similar to previous ones. RGLight outperforms other methods, especially as missing data increases. However, given higher probabilities of missing data, i.e., 60\%, both IGRL, and GAT-TSC perform worse than the Fixed time method, which might limit their usefulness.

Contrary to the Manhattan study, \emph{Greedy} performs worse than the Fixed time method. This result suggests that when the road network becomes more irregular as is the case for Luxembourg, \emph{Greedy} tends to fail. To confirm, we tested the \emph{Greedy} method on two synthetic networks with the same number of intersections, one with irregular road patterns (more similar to Luxemburg) and the second one laid out as a grid (similar to Manhattan). We confirm that \emph{Greedy} performs better on the latter.  


\begin{figure}[htbp]
\center
\includegraphics[width=9cm]{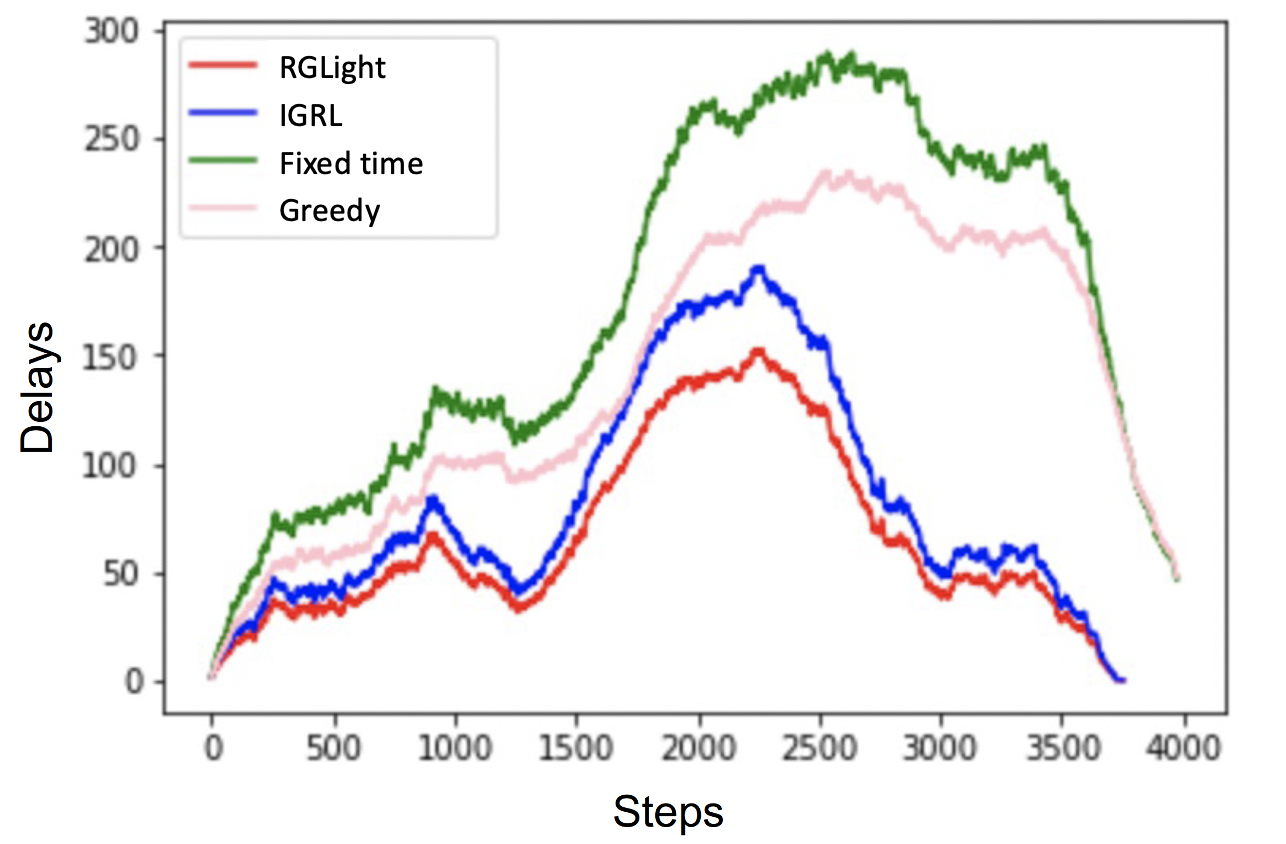}
\caption{Average delays evolution in Manhattan road network.}
\label{mhdelay}
\end{figure}

\begin{figure}[htbp]
\center
\includegraphics[width=9cm]{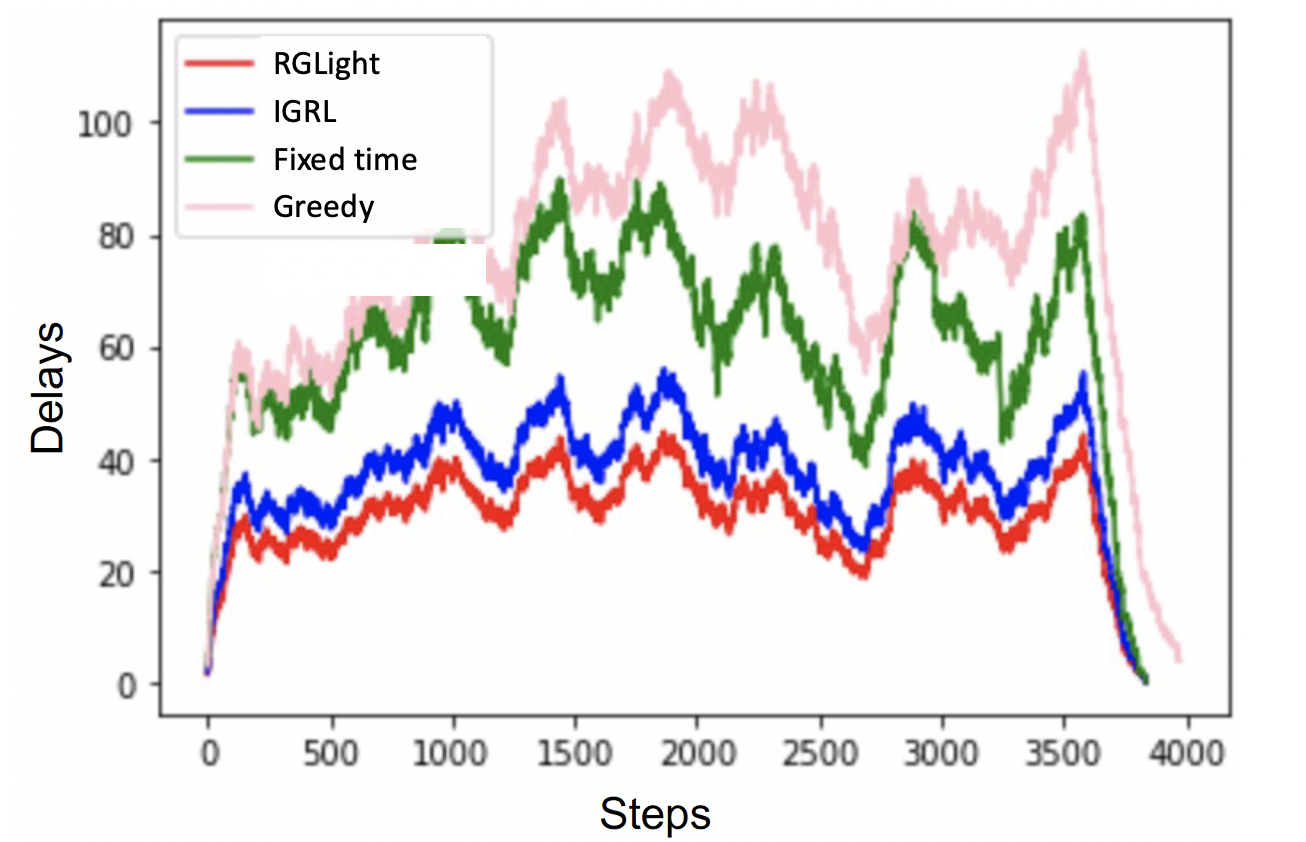}
\caption{Average delays evolution in Luxembourg road network.}
\label{lxdelay}
\end{figure}

To visualize the performance of the learned policy, we collect the average delays  per time step in two road networks. We select the best RL baseline and two transportation baselines. 
In Figure~\ref{mhdelay}, we see that RGLight better mitigates the effect of demand surge compared to other baselines. Moreover, from Figure~\ref{lxdelay}, faced with a more challenging demand evolution in the Luxembourg road network, RGLight also demonstrates the overall best robustness.

\begin{figure}[h]
\centering
\includegraphics[width=7cm]{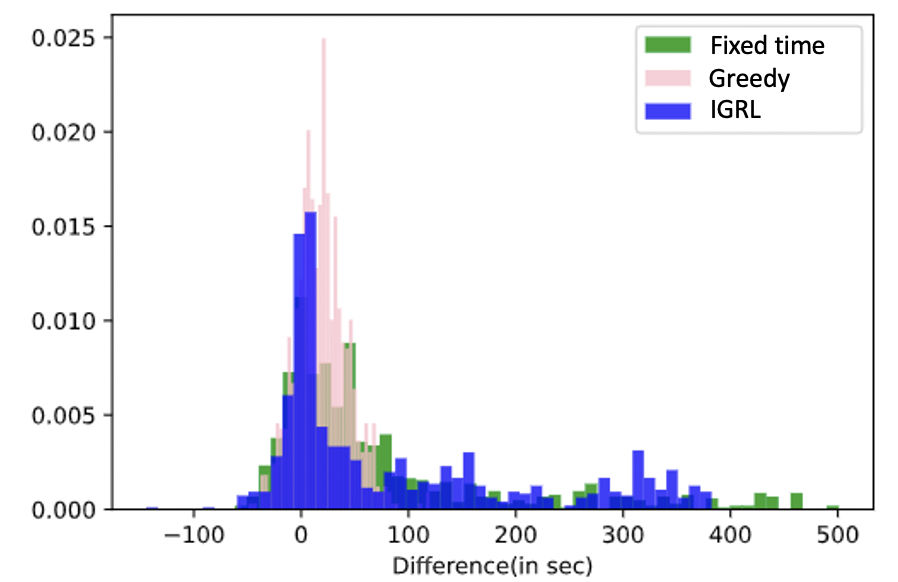}
\caption{Differences of paired trips travel time compared to RGLight. We report the difference between RGLight and the method (i.e. RGLight - method) and so numbers higher than 0 indicate the method being outperformed by RGLight. The y-axis is normalized.}
\label{tripdur}
\end{figure}

\subsection{Generalizability analysis}
\label{generalization ability}
Now we test more systematically the ability of the models to generalize to networks of different shapes and scales and under different traffic demands. 
This departs from most previous works~\citep{wei2019colight,zhang2020generalight,oroojlooy2020attendlight} that keep training and testing conditions similar.

We also introduce DGRL, a pure distributional baseline version of IGRL, obtained by setting $k=0$  in Equation~\ref{combineloss}. 

\begin{figure}[htbp]

\centering
\includegraphics[width=15cm]{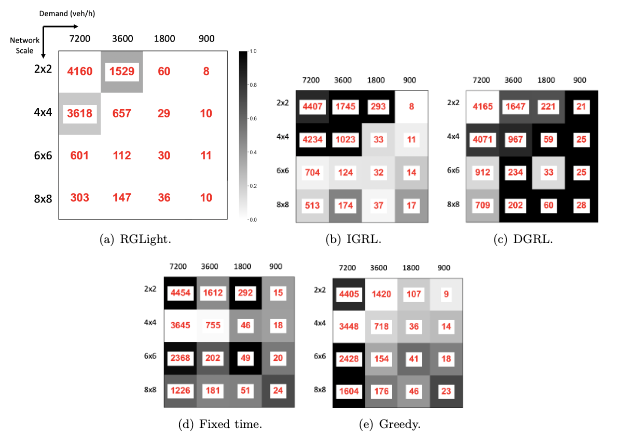}
\caption{ Comparison of generalizability using delay for different methods. The lateral direction on each sub-figure represents different traffic demands and the longitudinal direction represents different grid network scales (how many columns and rows are in the grid network). For example, in a scenario with a network scale of 4 and a demand of 0.5, we have a grid network with 4 columns and 4 rows and the arrival rate is 1/0.5=2 veh/seconds. The shading can only be compared across the methods by using the same scenario configuration (network scale and demand). For example, in a scenario with a network scale of 2 and a demand of 0.5, the Fixed time approach performs the worst so the color is darker compared to corresponding cells in other methods.  }
\label{gf}
\end{figure}

\begin{figure}[htbp]
\subsection{Interpretation of Learned Policies}
\label{switch rate}
\includegraphics[width=14cm]{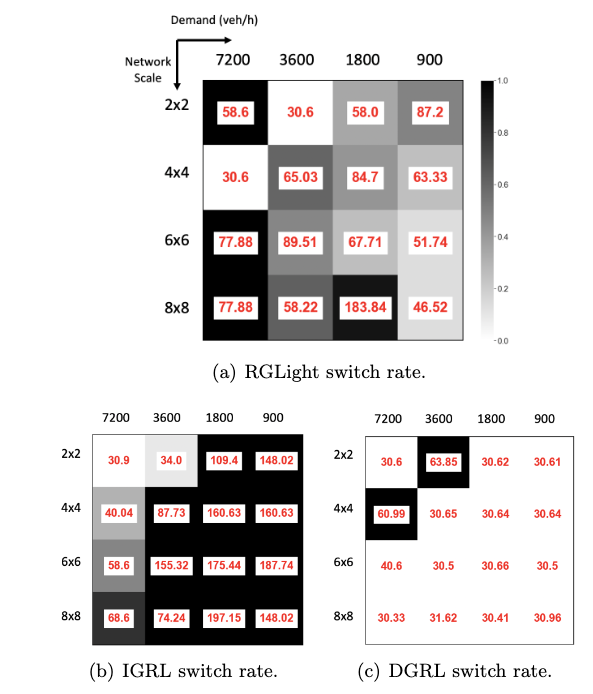}
\caption{ Comparison of switch rates for different methods.  We also use the same strategy to normalize the switch rate. Values closer to 1 indicate a higher switch rate. The numbers on each cell stand for the average switch rate multiplied by 1000.}
\label{sw-fig}
\end{figure}

We train models on \textit{irregular} synthetic networks with 2 to 6 intersections. The horizontal direction on each sub-figure in Figures~\ref{gf} and~\ref{sw-fig} represents different traffic demands (0.5, 1, 2, 4), and the vertical direction represents different grid network scales, that is, how many columns and rows in the grid network (4, 6, 8). In total, we test 16 different scenarios for each model to evaluate its generalizability. 

We use the average delay over the whole simulation process to evaluate model performance. Furthermore, we normalize the average delay of each method for readability: 



\begin{equation}
x_{i}'=\frac{x_i-x_{min}}{x_{max}-x_{min}} \times 10,000
\label{norm}
\end{equation}
where $x_i$ is the average delay calculated from method $i$,  $x_{max}$ and $x_{min}$ are the maximum and minimum delay calculated across all methods given the specific scenario.  Then we can use the normalized average delay to plot the colormap in Figure~\ref{gf}. The values of $x_{i}'$ range between 0 and 10,000 and smaller values indicate better performances.

Figure~\ref{gf} shows that all methods tend to perform worse for heavy-traffic regimes in small networks  (upper-left corner). This matches common knowledge about network traffic capacity~\citep{loder2019understanding}. We also find that the \emph{Greedy} baseline performs relatively well in small-scale networks but performs worse in large-scale networks.  We hypothesize it assumes that the downstream lanes have an unlimited capacity which makes it not very realistic in large-scale networks. As a result, we can see that the model's performance worsens when the network scale increases. This is similar to the finding in~\citet{wei2019colight}. On the other hand, we find that RL-based methods (i.e., IGRL and DGRL) are less sensitive to network scale change compared to the transportation method. This result demonstrates that RL methods can better generalize to different network structures  than standard transportation baselines.

We now focus on the reinforcement-learning methods. In the bottom right corner, IGRL performs better than DGRL, but DGRL performs better than IGRL in the upper-left corner (i.e., smaller network with higher demand). These results indicate  the weaker generalization ability of IGRL since its performance tends to decrease in test scenarios that are very different from the training scenarios (e.g., a small network under a heavy-traffic regime). We also find that DGRL performs better than IGRL in a small network with a heavy-traffic regime. We suspect that since the distributional approach uses a robust loss it might be less sensitive to outliers. 
However, in a normal traffic regime with a  larger network, DGRL performs worse than IGRL. 
These findings further motivate the  policy ensemble approach. Overall, we find that the RGLight method performs best across most scenarios. This result indicates that an ensemble of policies can boost generalizability.

   


To further analyze the characteristics of the policies learned by the RL methods, we examine the switch rates of IGRL, DGRL, and RGLight. Recall that the actions are binary and correspond to either switching to the next phase in a signal's program (action 1) or not switching (action 0). The switching rate is the ratio of signals that perform a phase switch (action 1) in a single timestep across all intersections. Using a similar matrix across network scale and demand as before, Figure~\ref{sw-fig} reports the average switch rate across methods. 

Comparing Figure~\ref{sw-fig} (b) and (c), we see that overall IGRL exhibits a higher switch rate compared to DGRL. In contrast, RGLight is often in-between IGRL and DGRL except when the demand is the highest (first column) and it switches more often than both. This seems to indicate that RGLight attains states that are different than the two other methods. 

We further discuss the scenario with a 2x2 network and a demand of 1800 veh/h. By considering Figure~\ref{gf} (a) and Figure~\ref{sw-fig} (a) together, we observe that RGLight does best. Further, its switch rate (58) is in-between IGRL's (109.4) and DGRL's (30.62). We provide a video demonstration of this simulation.\footnote{Simulation video link: \url{https://youtu.be/-n_LUbNjJUs}} In the video we notice that a policy that switches too often (IGRL) leads to a shock wave or gridlock. On the other hand, switching too slowly (DGRL) ends up preventing significant traffic from passing to allow less busy lanes to advance. RGLight seems to have found a good comprise. We believe it is worth further investigating how to design the signal phase and the action space  based on these types of results.



\section{Conclusions and Discussion}

Motivated by gaps in the current literature (Table~\ref{survey}), we propose RGLight, an RL approach that combines two reinforcement learning agents and that provides more generalizable and robust policies. Further, we conduct a series of experiments on two different real-world networks with real traffic demands and show that our method outperforms several state-of-the-art baselines.   

In future work, we plan to study the empirical and theoretical properties of RGLight to model multi-agent systems in other similar domains. Such general multi-agent settings include connected and automated vehicles environment \citep{wangmulti} and traffic junction environment \citep{liu2020multi}. 
As a second avenue, we will investigate combinations of RGLight (model-free) and model-based reinforcement learning that can both improve performance and also (training) data efficiency~\citep{schrittwieser2020mastering}.

\section*{Acknowledgment}

This research is supported by the Natural Sciences and Engineering Research Council (NSERC) of Canada, Mitacs Canada, the Canada Foundation for Innovation (CFI), and LC is supported by a Canada AI CIFAR Chair.

\bibliographystyle{elsarticle-harv}
\bibliography{reference}

\end{document}